\pdfoutput=1

\documentclass[11pt]{article}

\usepackage{acl}

\usepackage{times}
\usepackage{latexsym}

\usepackage[T1]{fontenc}

\usepackage[utf8]{inputenc}

\usepackage{microtype}
\usepackage{booktabs}
\usepackage{xcolor}
\usepackage{multirow}
\usepackage{graphicx}
\usepackage{subcaption}
\usepackage{mathbbol}
\usepackage{float}
\usepackage{amsmath,amsfonts,amssymb} 

\newcommand\blfootnote[1]{%
  \begingroup
  \renewcommand\thefootnote{}\footnote{#1}%
  \addtocounter{footnote}{-1}%
  \endgroup
}

\title{Hierarchical Attention Network for Explainable Depression Detection\\ on Twitter Aided by Metaphor Concept Mappings\textbf{}}

\author{Sooji Han$^*$, Rui Mao$^*$, and Erik Cambria\\
School of Computer Science and Engineering\\
Nanyang Technological University\\
  \texttt{\{sooji.han,rui.mao,cambria\}@ntu.edu.sg}}

\begin{document}
\maketitle
\begin{abstract}

\blfootnote{$*$ These authors contributed equally.}Automatic depression detection on Twitter can help individuals privately and conveniently understand their mental health status in the early stages before seeing mental health professionals. Most existing black-box-like deep learning methods for depression detection largely focused on improving classification performance. However, explaining model decisions is imperative in health research because decision-making can often be high-stakes and life-and-death. Reliable automatic diagnosis of mental health problems including depression should be supported by credible explanations justifying models' predictions. In this work, we propose a novel explainable model for depression detection on Twitter. It comprises a novel encoder combining hierarchical attention mechanisms and feed-forward neural networks. To support psycholinguistic studies, our model leverages metaphorical concept mappings as input. Thus, it not only detects depressed individuals, but also identifies features of such users' tweets and associated metaphor concept mappings. 

\end{abstract}

\section{Introduction}\label{intro}

Depression is a serious health and social issue that afflicts many individuals in modern society and its prevalence is predicted to increase globally. People with depression are likely to express their feelings and mental states over their social media before seeing health professionals~\cite{guntuku2017detecting, ansari2022ensemble}. An automatic, efficient approach for depression identification is imperative to recommend adequate treatment, achieving remission and preventing relapse. Recent studies on automatic depression detection on social media~\cite{gui2019cooperative,lin2020sensemood,ji2021mentalbert,zogan2021depressionnet} have largely focused on achieving higher detection accuracy. However, it is impossible to explain and interpret those black-box models that rely on state-of-the-art (SOTA) deep learning techniques. The recent development of explainable AI emphasizes that it is crucial for health professionals to fully comprehend, monitor and trust the AI decision-making mechanisms.

People suffering from depression often use metaphors to describe their emotions and the experience of living with mental illness~\cite{coll2021metaphors,roystonn2021analysis}. In psychotherapy, metaphors are a pivotal tool for helping people with depression better understand themselves and their problems and facilitating effective communication between therapists and patients~\cite{kopp2013metaphor,siegelman1993metaphor}. This is because metaphorical expressions implicitly reflect people's different ways of understanding the same target. Metaphor is not only a linguistic phenomenon, but also a reflection of cognitive mappings of source and target concepts~\cite{lakoff1980metaphors}. Analyzing metaphor concept mappings (MCMs) helps us understand the inner world of people with depression. Metaphoric expressions associated with depression have been widely studied in psychology, particularly as a form of case studies~\cite{roystonn2021analysis,coll2021metaphors}. To the best of our knowledge, however, there has not been an automatic method that leverages MCM features extracted from a large corpus for depression detection. We are motivated to bridge the gap and offer better insights into automatic depression detection on social media and conceptual metaphor understanding. Furthermore, we argue that psychological and psycholinguistic research communities can benefit from automated, explainable tools for studying the relationship between depression and metaphors.

In this work, we propose an explainable framework for depression detection on Twitter, called \textit{Hierarchical Attention Network (HAN)}. We propose a novel attention-based encoder which allows HAN to learn important inputs for user-level binary classification (i.e., depressed and non-depressed users). To further improve the interpretability of depression detection, we introduce MCMs as an additional feature into the model. Health professionals and potential patients can use learned features (i.e., characteristics of depressive tweets and MCMs) as justification. We evaluate our model on a publicly available Twitter depression detection dataset~\cite{shen2017depression} and show that HAN achieves the SOTA performance. It outperforms the strongest baseline~\cite{zogan2021depressionnet} by increasing an F1 score by 6.0\% on average. Additionally, our newly proposed encoder outperforms several classical encoders. In particular, HAN improves LSTM~\cite{hochreiter1997long} (the most competitive benchmark encoder for our task) by 1.9\% on a validation set with a quarter of the number of parameters of LSTM. Finally, we visualize and analyze examples of attention weights learned by HAN to demonstrate its explainability.

The main contributions of this work can be summarized as follows: (1)  We propose an explainable model for depression detection on Twitter. Unlike most SOTA methods employing attention mechanisms at word level~\cite{vaswani2017attention,liu2021domain}, our model employs context-level attention mechanisms to identify the relative importance of certain tweets and metaphors, which is crucial for filtering out less significant information in the final representation of contexts and justifying the outputs of the model. (2) We introduce MCMs as a feature to improve explainability and performance. This also helps a better understanding of the cognition of depressive individuals. (3) We demonstrate that HAN achieves outstanding performance and produces accurate and explainable results with a smaller number of training parameters than classical encoders via extensive experiments.

\section{Related Work} \label{related_work}
Traditional studies on depression focus on social, psychological and biological factors, which are not often readily available. This paper mainly focuses on social media texts and machine-learning-based depression detection. Several studies in psychology have reported that conceptual metaphors are used to express and understand the experience of depression, but they are often used unconsciously and pass unnoticed. Research into metaphors can help better understand individuals with depression.

\noindent \textbf{Depression Detection on Social Media.}~\citet{zogan2021depressionnet} proposed a model combining CNNs~\citep{lecun1989backpropagation} and BiGRUs~\citep{deng2019sequence} for learning users' behavior and textual contents. For user behavior modeling, manually curated features, which are associated with emotions, domains, topics and social media metadata, were employed. Some research exploited sentiment analysis techniques for depression detection.~\citet{rao2020mgl} proposed a hierarchical architecture leveraging gated units and CNNs to learn textual contents of social media posts and users' emotional states expressed in posts.~\citet{aragon2021detecting} proposed an emotion-aware SVM-based model which learns emotional dynamics expressed in social media posts.~\citet{chiong2021combining} proposed 90 features, based on sentiment lexicons and textual contents and used them as input to depression detection classifiers. A recent trend is to exploit multimodal learning frameworks for depression detection~\cite{gui2019cooperative,chiu2021multimodal,lin2020sensemood,yang2018integrating}.~\citet{gui2019cooperative} proposed a multimodal multi-agent reinforcement learning model incorporating BiGRU and VGGNet~\citep{simonyan2014very} to learn texts and images posted by users on Twitter, respectively.~\citet{chiu2021multimodal} proposed a multimodal BiLSTM-based~\cite{schuster1997bidirectional} architecture jointly learning texts, images and temporal behaviors (i.e., time intervals between posts) on Instagram.~\citet{lin2020sensemood} proposed a multimodal model comprising a CNN and a BERT. It jointly learns representations of images and texts and fused them using a low-rank multimodal fusion method.~\citet{zhang2021mam} proposed a model combining BiLSTM and CNN based on metaphor features and text. However, their metaphor features are shallow, e.g., Part-of-Speech (PoS) tags and the number of metaphors.

\noindent \textbf{Metaphor Understanding.} Traditional metaphor studies on depression were mainly based on qualitative analysis and case studies~\cite{roystonn2021analysis,coll2021metaphors}. This is because of a lack of automatic tools that help psycholinguistic researchers parse and analyze metaphorical expressions from large corpora. Recently, automatic metaphor processing has achieved significant developments. Metaphors can be identified with sequence-tagging-based models~\citep{mao2019end,choi2021melbert,chen2021metaphor,mao2021bridging}. Then, the identified metaphors can be interpreted by linguistic meanings~\citep{bollegala2013metaphor,mao2018word,mao2021interpreting} or concept mappings~\citep{mason2004cormet,shutova2017multilingual,ge2022explainable}. Linguistic metaphor interpretation focuses on paraphrasing metaphors into their literal counterparts. For example, \citet{mao2022metapro} proposed a system for metaphor identification and interpretation, called MetaPro. It can be used as a text pre-processing technique to achieve metaphor paraphrases from end to end. Thus, NLP techniques for downstream tasks, such as sentiment analysis \citep{mao2022metapro} or machine translation \citep{mao2018word}, can achieve better performance on the effectiveness of metaphor paraphrasing. However, \citet{lakoff1980metaphors} argued that metaphor is not only a linguistic phenomenon, but also a reflection of humans' cognition. Given ``this is the \textit{core}\footnote{A metaphor is in italics.} of the matter'', \textit{core} implies ``importance'' (target) is ``interiority'' (source)~\citep{lakoff1991master}. Thus, one can analyze the inner world of depressed people from their metaphoric expressions and the associated concept mappings, e.g., \textsc{importance is interiority}. 

In this paper, we identify several limitations of existing works on depression detection. Existing methods have largely focused on improving classification performance by using advanced encoders, features and deep architectures, while leaving model outputs inexplicable. The majority of SOTA methods are limited to textual contents of posts or rely on shallow features based on social media metadata. To our best knowledge, it is the first work incorporating MCMs into machine-learning-based depression detection on social media. Additionally, our model comprises context-level explainable encoders while word-level attention mechanisms have been widely employed in SOTA methods. This helps better understand how certain tweets and MCMs are used by depressed individuals, thereby justifying model predictions. 

\section{Methodology}

\subsection{Problem statement}

In our task, a user ($u_k$) is represented as a set of tweets ($\mathbb{X}_k$) and a set of the associated MCMs ($\mathbb{M}_k$) in the tweets. Therefore, a set of users is denoted by $\mathbb{U}=\{u_1,\cdots,u_i\}$, where each user $u_k=[\mathbb{X}_k, \mathbb{M}_k]$.  A set of a user's \textbf{tweet contents} is denoted by $\mathbb{X}_k=\{x_{k,1},\cdots,x_{k,n}\}$ which contains $n$ tweets. $x$ is a tweet represented as a sequence of words. A set of a user's \textbf{MCMs} is denoted by $\mathbb{M}_k=\{m_{k,1},\cdots,m_{k,s}\}$ which contains $s$ mappings. $m$ is an MCM that is represented as a sequence of ``\textsc{a is b}'',  where \textsc{a} is a target concept, \textsc{b} is a source concept, and \textsc{is} a relation mapping \textsc{a} to \textsc{b}. An example is ``\textsc{importance is interiority}''. The task is to predict the most probable label ($\hat{y}_k$) for a user $u_k$, where $\hat{y}_k\in\{0,1\}$. $\hat{y}_k=1$ if $u_k$ is a depressed user, $\hat{y}_k=0$ otherwise. $y_k$ denotes a ground truth label.

\subsection{Model Architecture}

\begin{figure*}[t!]
    \centering
    \includegraphics[width=0.88\linewidth]{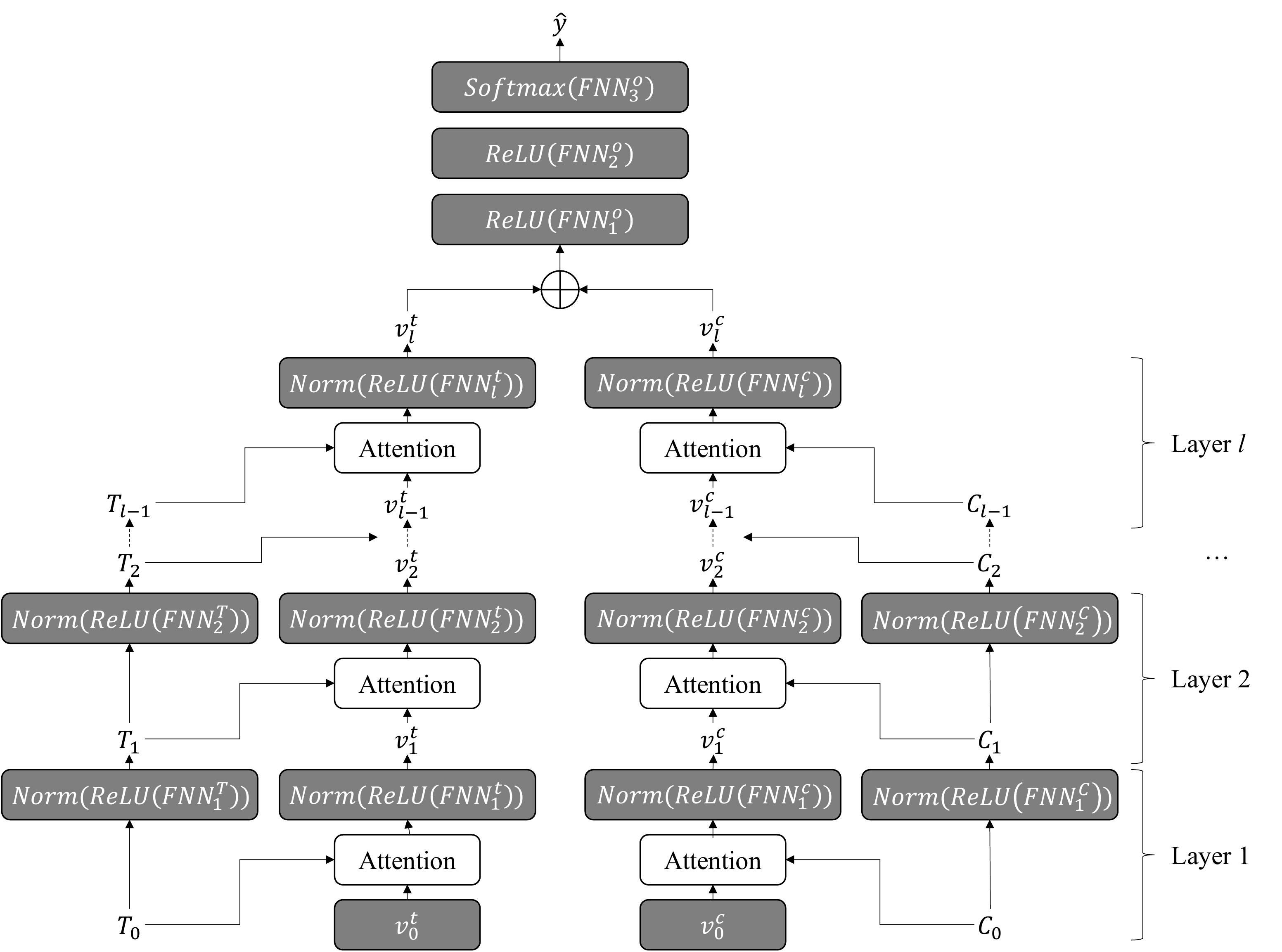}
    \caption{Hierarchical Attention Network. Grey boxes denote computational layers with trainable parameters. Plain text denotes input and output. $T$ and $C$ denote tweet and MCM embeddings, respectively.}
    \label{fg:framework}
\end{figure*} 

The overall architecture of HAN is shown in Figure~\ref{fg:framework}. Given a set of tweets ($\mathbb{X}$) of a user, we first obtain embeddings of all tweets in $\mathbb{X}$ using a pre-trained language model. BERT-base-uncased~\cite{devlin2018bert} is used to be in line with our strongest baseline~\cite{zogan2021depressionnet}. Special tokens [CLS] and [SEP] are added at the beginning and at the end of each tweet $x_\epsilon$, respectively. The padded sequence ``[CLS] $x_\epsilon$ [SEP]'' is fed into BERT. The vector representation at the [CLS] position of BERT output is used as the embedding of $x_\epsilon$. We obtain an embedding matrix of all tweets in $\mathbb{X}$, denoted by $T_0$. Formally,
\begin{equation}
    T_0 = BERT(\mathbb{X}).
\end{equation}
Similarly, we obtain an embedding matrix of all MCMs in $\mathbb{M}$, denoted by $C_0$. Formally,
\begin{equation}
    C_0 = BERT(\mathbb{M}).
\end{equation}
Details about the acquisition of $\mathbb{M}$ are described in Section~\ref{sec:Concept mapping acquisition}.

HAN consists of $l$ attention-based encoder layers. The $i^{th}$ encoder layer is defined as $HAN_i(\cdot)$ (see Section~\ref{sec:Hierarchical attention network encoder} for details), where $i\in\{1,2,\cdots l\}$. Given a query vector and a key matrix, $HAN_i(\cdot)$ yields an updated query vector and key matrix. Thus, given the query vector ($v_{i-1}^{t}$) of tweets ($t$) and the tweet embedding (key) matrix $T_{i-1}$ in the ${(i-1)}^{th}$ layer, the updated $v_{i}^{t}$ and $T_{i}$ are given by 
\begin{equation}
\label{eq:T_i}
    v^t_i, T_i = HAN^t_i(v_{i-1}^{t}, T_{i-1}).
\end{equation}
Similarly, given the query vector ($v_{i-1}^{c}$) of MCM ($c$) and the MCM embedding (key) matrix $C_{i-1}$ in the ${(i-1)}^{th}$ layer, the updated $v_{i}^{c}$ and $C_{i}$ are given by
\begin{equation}
\label{eq:C_i}
    v^c_i, C_i = HAN^c_i(v_{i-1}^{c}, C_{i-1}).
\end{equation}
For the first layers of the first training step, the inputs $v_0^{t}$ and $v_0^{c}$ (trainable parameters) are randomly initialized. For the other training steps, $v_0^{t}$ and $v_0^{c}$ are values learned in the previous step.

Next, the output of the last layer of the tweet encoder ($v_l^{t}$) and that of the MCM encoder ($v_l^{c}$) are concatenated ($\oplus$). The concatenated representation is fed to three feed-forward neural networks (FNNs), denoted by $FNN^{o}(\cdot)$. The first two $FNN^{o}(\cdot)$ are activated by ReLU ($ReLU(\cdot)$)~\citep{agarap2018deep}. The last FNN, $FNN_{3}^{o}(\cdot)$, is activated by the Softmax. We do not change the size of the hidden state given by the outputs of the first two FNN layers. The last FNN layer projects the hidden state into a vector of the label size. Then, the probability of a predicted label ($\hat{y}$) is given by
\begin{align}
\label{eq:concatenate}
    h &= ReLU(FNN^{o}(v_{l}^{t} \oplus v_{l}^{c}))_{\times 2}\\
    \hat{y} &= Softmax(FNN_{3}^{o}(h))),
\end{align}
where $h$ is the hidden state after the first two ($\times 2$) FNNs. Cross-entropy loss is used to optimize the parameters in the model and is given by
\begin{equation}
    \mathcal{L} = CrossEntropy(\hat{y}, y).
\end{equation}

\subsection{Concept mapping acquisition}
\label{sec:Concept mapping acquisition}

Concept mapping acquisition process consists of three components: \textbf{a) metaphor identification} ($MI(\cdot)$)~\cite{mao2021bridging}, \textbf{b) metaphor paraphrasing} ($MP(\cdot)$)~\cite{mao2021interpreting} and \textbf{c) concept mapping generation} ($CG(\cdot)$)~\cite{ge2022explainable}. These algorithms are used because they enable the end-to-end acquisition of MCM features without pre-processing and domain-specific knowledge. Here, we briefly introduce their algorithms, inputs and outputs. For the details, please refer to the original papers.

Given a tweet ($\mathbb{x}_\epsilon$) comprising $g$ tokens $\tau$, i.e., $\mathbb{x}_\epsilon=\{\tau_{\epsilon,1}, \tau_{\epsilon,2}, \cdots, \tau_{\epsilon,g}\}$. The \textbf{metaphor identification} module (MI) is a multi-task-learning-based sequence tagging model, yielding a metaphor label sequence ($r_\epsilon$) and a PoS label sequence ($\rho_\epsilon$) defined by
\begin{equation}
    r_\epsilon, \rho_\epsilon = MI(\mathbb{x}_\epsilon),
\end{equation}
where $r_\epsilon = \{r_{\epsilon,1},r_{\epsilon,2},\cdots,r_{\epsilon,g}\}$ and $\rho_\epsilon = \{\rho_{\epsilon,1},\rho_{\epsilon,2},$ $\cdots,\rho_{\epsilon,g}\}$. $r_{\epsilon,j}\in \{\rm{metaphor}, \rm{literal}\}$ and $\rho_{\epsilon,j}$ is a Universal-Dependency-scheme-based PoS label, where $j\in\{1,2,\cdots,g\}$ denotes the index of a token in $\mathbb{x}_\epsilon$. To boost model performance,~\citet{mao2021bridging} proposed a Gated Bridging Mechanism for soft-parameter sharing between the metaphor identification and PoS tagging tasks. 

Next, given an identified metaphoric open-class word\footnote{Closed-class words are not paraphrased because they do not convey much semantic information in their context.} $\tau_{\epsilon,j}$ (i.e., one of verbs, nouns, adjectives and adverbs), the \textbf{metaphor paraphrasing} module first lemmatizes $\tau_{\epsilon,j}$ as $\tau_{\epsilon,j}^{\iota}$. Then, a pre-trained language model is used to select the best fit word~($\omega_{\epsilon,j}$) from a candidate set that consists of hypernyms and synonyms of $\tau_{\epsilon,j}^{\iota}$ in WordNet~\cite{fellbaum2005wordnet} and their inflections with the same PoS. The best fit word denotes a candidate word that appears in the context and has the highest probability. A probability is given by a masked word prediction of the pre-trained language model, which has been widely used in prompt-based zero-short learning tasks \citep{mao2022biases}. The best fit word $\omega_{\epsilon,j}$ is lemmatized as $\omega_{\epsilon,j}^{\iota}$, which is considered the lemma of the paraphrased metaphor $\tau_{\epsilon,j}$ in the context of $x_\epsilon$. The above process is defined by
\begin{equation}
    \omega_{\epsilon,j}^{\iota} = MP(\tau_{\epsilon,j}, \rho_{\epsilon,j}).
\end{equation}

Finally, the \textbf{concept mapping generation} module abstracts the source concept (\textsc{a}$_{\epsilon,j}$) from $\tau_{\epsilon,j}^{\iota}$ and the target concept (\textsc{b}$_{\epsilon,j}$) from $\omega_{\epsilon,j}^{\iota}$. Formally,
\begin{align}
    \textrm{\textsc{a}}_{\epsilon,j} &= CG(\tau_{\epsilon,j}^{\iota}),\\ \textrm{\textsc{b}}_{\epsilon,j} &= CG(\omega_{\epsilon,j}^{\iota}).
\end{align}
$CG(\cdot)$ is a knee algorithm~\citep{satopaa2011finding} and a WordNet-based conceptualization method, proposed by~\citet{ge2022explainable}. It abstracts a word into a concept by looking up a hypernym that can cover the major senses of a word. After obtaining \textsc{a}$_{\epsilon,j}$ and \textsc{b}$_{\epsilon,j}$, the concept mapping is defined as 
\begin{equation}
\label{eq:concept mapping}
    \rm{MCM}_{\epsilon,\textit{j}} = \textsc{b}_{\epsilon,\textit{j}}~\textsc{is}~\textsc{a}_{\epsilon,\textit{j}}.
\end{equation}

\citet{ge2022explainable} argued that~\citet{lakoff1991master} summarized concept mappings with different patterns due to the subjectivity of annotators. We take the concept mappings given by Eq.~\ref{eq:concept mapping}, which follows one of the concept mapping principles of~\citet{lakoff1991master} (see Section~\ref{related_work}). We obtain all concept mappings in $\mathbb{x}_\epsilon$. If no metaphor is detected in $\mathbb{x}_\epsilon$, concept mapping is none for such a tweet. All concept mappings from all tweets of each user are collected, forming an MCM feature set ($\mathbb{M}$) for depression detection.

\subsection{Hierarchical attention network encoder}
\label{sec:Hierarchical attention network encoder}

The HAN encoder ($HAN(\cdot)$) is based on scaled dot-product attention and FNNs. Attention mechanisms enable the model to identify input features (i.e., tweets and MCMs) that are highly significant and useful for depression detection, thereby enhancing model explainability. FNNs allow feature embeddings to better fit the task via multiple non-linear projections. Unlike self-attention~\cite{vaswani2017attention}, the feature information of our encoder is not shared with each other within each feature set, i.e., $T_i$ and $C_i$ in Eqs.~\ref{eq:T_i} and \ref{eq:C_i}. Thus, features fed to the last encoder layer (i.e., $T_{l-1}$ and $C_{l-1}$) represent the features of individual inputs even after several non-linear projections. Important features are learned by query vectors. These are the main differences between our explainable encoder and other classical black-box-like encoders, e.g., LSTM, BiLSTM, GRU~\cite{cho2014properties}, BiGRU and Transformer~\cite{vaswani2017attention}, hidden states of which cannot be easily disentangled after encoding.

Given a query vector ($q_{i-1}\in\mathbb{R}^{1\times d}$) and a key matrix ($K_{i-1}\in\mathbb{R}^{o\times d}$) in the ${(i-1)}^{th}$ layer, where $d$ denotes an embedding size and $o$ denotes the number of input features, attention weights ($w_{i}\in\mathbb{R}^{1\times o}$) in the $i^{th}$ layer are given by
\begin{equation}
\label{eq:attention weight}
    w_{i} = Softmax\left(\frac{q_{i-1}\otimes K_{i-1}^{\intercal}}{\sqrt{d}}\right),
\end{equation}
where $\otimes$ denotes matrix product. The query vector ($q_{i}\in\mathbb{R}^{1\times d}$) in the $i^{th}$ layer is given by the weighted sum of the vectors in $K_{i-1}$ and a non-linear projection. Formally,
\begin{equation}
\label{eq:attention_sum}
    q_i\!=\!LN(ReLU(FNN^{query}_i(w_{i}\otimes\!K_{i-1}))),
\end{equation}
where $LN(\cdot)$ denotes layer normalization~\citep{ba2016layer}. The key matrix ($K_{i}\in\mathbb{R}^{o\times d}$) in the $i^{th}$ layer is defined by
\begin{equation}
\label{eq:key_projection}
    K_i = LN(ReLU(FNN^{key}_i(K_{i-1}))).
\end{equation}

The input and output of the HAN encoder have the same size. For the tweet content encoder, $q, K$ and $o$ denote a tweet query vector ($v^{t}$), a tweet embedding matrix ($T$) and the number of tweets ($n$), respectively. For the MCM encoder, $q, K$, and $o$ denote an MCM query vector ($v^{c}$), an MCM embedding matrix ($C$) and the number of MCMs ($s$), respectively. We use the attention weights from the last (the $l^{th}$) encoder layer as the final representation of tweet contents and MCMs. $w_{l}$ shows important inputs that have higher attention weights for the depression status prediction of a user. Analysis results of attention weights to demonstrate model explainability are described in Section~\ref{sec:Explainability demonstration}.



\section{Experiments}
\subsection{Datasets and pre-processing}\label{data_set}

\begin{table}
\centering
\small
\begin{tabular}{ccc|cc} 
\toprule
\multirow{2}{*}{\textbf{Dataset}} & \multicolumn{2}{c|}{\textbf{Total \# of tweets }} & \multicolumn{2}{c}{\begin{tabular}[c]{@{}c@{}}\textbf{Mean \# of tweets}\\\textbf{ per user }\end{tabular}}  \\ 
\cline{2-5}
 & \textbf{Positive} & \textbf{Negative} & \textbf{Positive} & \textbf{Negative}\\ 
\hline
D1  & 156,013  & 153,328  & 72 & 75\\
D2  & 151,538  & 119,188  & 71 & 58\\
D3  & 142,057  & 118,611  & 66 & 58\\
D4  & 143,725  & 124,925  & 66 & 61\\
D5  & 148,039  & 134,700  & 69 & 66\\
\bottomrule
\end{tabular}
\caption{\label{data_stats}Statistics of the five randomly sampled datasets. The number of positive users and that of negative users are 2,159 and 2,049 for all the datasets.}
\end{table}


Table~\ref{data_stats} presents the statistics of the datasets used in our experiments. We use a publicly available Twitter dataset, called MDL~\citep{shen2017depression}, which was designed for depression detection. In this dataset, Twitter users, who have posted tweets containing pre-defined patterns (i.e., I'm/I was/I am/I've been diagnosed depression), were labeled as depressive (i.e., positive). Those who never posted any tweet containing the term ``depress'' were labeled as users not suffering from depression (i.e., negative). Due to the updates of MDL over time, its statistics varies across existing works which used it for their experiments~\citep{gui2019cooperative,lin2020sensemood,zogan2021depressionnet}. We argue that the model of~\citet{zogan2021depressionnet} is the most comparable to our model as regards architecture and features, e.g., employing two independent encoders to encode textual features from multiple sources. To make our results comparable with the work of \citet{zogan2021depressionnet}, 2,159 positive and 2,049 negative users are randomly sampled from the latest version of MDL. For a fair comparison, we generate five datasets with randomly selected users. 60\%, 20\% and 20\% of the full dataset are used for train, validation and test sets, respectively, which results in 2,524 users in a train set and 842 users in each of validation and test sets. We exclude tweets with less than 4 tokens because they are less informative for depression detection and analysis. URLs, mentions and emojis are removed from tweets because they are likely to introduce noise to the classification task~\citep{gao2020rpdnn}.

\subsection{Baselines}

We compare our model with three depression detection baselines. 

\noindent $\bullet$~\citet{gui2019cooperative}: A reinforcement-learning-based model based on cooperative multi-agent policy gradients. Tweet texts and images are encoded using GRUs and VGGNets, respectively. 

\noindent $\bullet$~\citet{lin2020sensemood}: A model comprising a CNN and a BERT for learning images and texts, respectively. The final representation of inputs is obtained via low-rank multimodal fusion.

\noindent $\bullet$~\citet{zogan2021depressionnet}: A model jointly learning tweet texts and user behavior using CNNs and BiGRUs. BERT-base and BART-large are used for tweet text modeling.

We do not benchmark the work by~\citet{zhang2021mam} because their model was designed for classifying different types of mental disorders. Besides, their model (i.e., CNNs+BiLSTMs) is similar to the architecture proposed by~\citet{zogan2021depressionnet}.

\begin{table}[t!]
\centering
\small
\begin{tabular}{p{\dimexpr 0.42\linewidth-2.0\tabcolsep}|
p{\dimexpr 0.13\linewidth-2.0\tabcolsep}
p{\dimexpr 0.13\linewidth-2.0\tabcolsep}
p{\dimexpr 0.13\linewidth-2.0\tabcolsep}
p{\dimexpr 0.13\linewidth-2.0\tabcolsep}} 
\toprule
\textbf{Model} & \textbf{P} & \textbf{R} & \textbf{F1} & \textbf{Acc.}  \\ 
\hline
\citet{gui2019cooperative} & 0.900  & 0.901  & 0.900 & 0.900 \\
\citet{lin2020sensemood} & 0.903  & 0.870  & 0.886 & 0.884 \\
\citet{zogan2021depressionnet} & 0.909  & 0.904  & 0.912 & 0.901 \\
\hline
HAN$_{\rm ours}$-Avg$_{\rm D1-D5}$  & 0.975  & 0.969  & \bf 0.972 & \bf 0.971 \\ 
\hline
\hline
D1 & 0.981  & 0.965  & 0.973 & 0.973 \\
D2 & 0.988  & 0.956  & 0.972 & 0.971 \\
D3 & 0.972  & 0.972  & 0.972 & 0.971 \\
D4 & 0.968  & 0.970  & 0.969 & 0.968 \\
D5 & 0.964  & 0.981  & 0.972 & 0.971 \\
\bottomrule
\end{tabular}
\caption{\label{tab:results} Depression detection results. Our model result is averaged over the five testing sets (D1-D5).}
\end{table}

\subsection{Setups}

We employ two HAN encoder layers (i.e., $l=2$). The maximum input length (i.e., the maximum numbers of tweets and MCMs per user) is set to 200. The batch size (i.e., the number of users per batch) is 64. Dropout rates for query vectors and key matrices are set to 0.2. The learning rate and weight decay of the Adam optimizer~\citep{kingma2015adam} are set to 1e-4 and 1e-5, respectively. BERT-base-uncased is used to obtain tweet and MCM embeddings. The model is trained with a GeForce GTX 1080 Ti GPU with CUDA 9.2~\citep{cuda9} and PyTorch 1.7.1~\citep{paszke2019pytorch}. Following our baselines, four performance metrics are adopted in our experiments: accuracy (Acc.), precision (P), recall (R) and F1 score (F1). P, R and F1 are computed with respect to the positive class, i.e., depressive. Overall performance is the average of results achieved with the five datasets.

\begin{table*}
\centering
\small
\begin{tabular}{
p{\dimexpr 0.126\linewidth-1.2\tabcolsep}|
p{\dimexpr 0.06\linewidth-1.2\tabcolsep}
p{\dimexpr 0.06\linewidth-1.2\tabcolsep}
p{\dimexpr 0.06\linewidth-1.2\tabcolsep}
p{\dimexpr 0.06\linewidth-1.2\tabcolsep}
p{\dimexpr 0.06\linewidth-1.2\tabcolsep}
p{\dimexpr 0.06\linewidth-1.2\tabcolsep}|
p{\dimexpr 0.06\linewidth-1.2\tabcolsep}
p{\dimexpr 0.06\linewidth-1.2\tabcolsep}
p{\dimexpr 0.06\linewidth-1.2\tabcolsep}
p{\dimexpr 0.06\linewidth-1.2\tabcolsep}
p{\dimexpr 0.06\linewidth-1.2\tabcolsep}
p{\dimexpr 0.06\linewidth-1.2\tabcolsep}
} 
\toprule
\multirow{2}{*}{\textbf{Model}} & \multicolumn{6}{c|}{\textbf{F1 on MDL-validation}}     & \multicolumn{6}{c}{\textbf{F1 on IMDL-validation}}      \\
    & \textbf{D1} & \textbf{D2} & \textbf{D3} & \textbf{D4} & \textbf{D5} & \textbf{Avg} & \textbf{D1} & \textbf{D2} & \textbf{D3} & \textbf{D4} & \textbf{D5} & \textbf{Avg}  \\ \hline
HAN      & 0.985 & 0.960 & 0.971 & 0.976 & 0.975 & \textbf{0.973}  & 0.939 & 0.911 & 0.933 & 0.927 & 0.931 & \textbf{0.928} \\
HAN-MCM   & 0.972 &  0.947  & 0.963 & 0.967 & 0.962 & \textbf{0.962} & 0.914 & 0.897 & 0.914  & 0.905  &  0.918 &  \textbf{0.909} \\ \hline
$\Delta$   & 0.013 &  0.013  &  0.008 & 0.009  & 0.013 & \textbf{0.011} & 0.025 & 0.014 & 0.019 & 0.022 & 0.014 & \textbf{0.019} \\
\bottomrule
\end{tabular}
\caption{\label{ablation} Ablation study results on validation sets, measured by F1 score. $\Delta$ is defined by $F1_{\text{HAN}}-F1_{\text{HAN-MCM}}$.}
\end{table*}

\section{Results}

\subsection{Classification performance}

As shown in Table~\ref{tab:results}, our proposed model advances SOTA performance in terms of all of the performance metrics for all of the five datasets. HAN achieves an average F1 score of 97.2\% and an accuracy of 97.1\%. The comparison results show that HAN yields the increases of 6.0\% and 7.0\% in F1 score and accuracy over the strongest baseline model~\citep{zogan2021depressionnet}, respectively. We observe that performance is almost identical for different randomly sampled datasets (i.e., D1-5), which shows that HAN is robust to different characteristics of users on Twitter. 

\subsection{Ablation study}

A set of exploratory experiments is conducted to study the relative contribution of MCMs in our model. To this end, we generate a variation of MDL, called Implicit Twitter Depression Data (IMDL), by removing explicit linguistic cues for depression (i.e., ``I'm/I was/I am/I've been diagnosed depression'' and words containing ``depress'', ``diagnos'', ``anxiety'', ``bipolar'' and ``disorder'') from all tweets. IMDL makes the task more challenging. 

We evaluate our full model (HAN) and HAN without the MCM encoder, called ``HAN-MCM'', using both MDL and IMDL. Table~\ref{ablation} shows average F1 scores achieved with the validation sets. Overall, HAN and HAN-MCM achieve higher performance on MDL than on IMDL in terms of F1 score. This verifies our hypothesis that the removal of explicit linguistic cues for depression from tweets makes the task more difficult. The ablation study of the internal baseline model (i.e., ``HAN-MCM'') proves that MCMs can provide additional information effective in identifying depressive users. 

For MDL, HAN outperforms HAN-MCM by 1.1\%. It is worth noting that the performance difference is slightly larger for IMDL (1.9\%), which indicates that MCMs can provide effective complementary evidence when no explicit cues for depression exist in tweets. The above experiments on both datasets demonstrate the usefulness of understanding MCMs in identifying depressive users. 

\begin{table}[t!]
\centering
\small
\begin{tabular}{
p{\dimexpr 0.4\linewidth-1.2\tabcolsep}
p{\dimexpr 0.1\linewidth-1.2\tabcolsep}
p{\dimexpr 0.1\linewidth-1.2\tabcolsep}
p{\dimexpr 0.1\linewidth-1.2\tabcolsep}
p{\dimexpr 0.1\linewidth-1.2\tabcolsep}} 
\toprule
\textbf{\# of HAN layers} & \textbf{1} & \textbf{2} & \textbf{4} & \textbf{8}  \\ 
\hline
MDL D1 validation & 0.542 & \textbf{0.985} & 0.983 & 0.979 \\
\bottomrule
\end{tabular}
\caption{F1 scores for different numbers of encoder layers.}
\label{tb: Layer analysis}
\end{table}

\subsection{Hyperparameter analysis}

The major hyperparameter of HAN is the number of encoder layers (i.e., $l$). We experiment with different values (i.e., 1, 2, 4 and 8) for the MDL D1 validation set. Table~\ref{tb: Layer analysis} shows that the best F1 score is achieved when $l$ is set to 2. Using more than two layers does not reap benefits of an increase in model performance.

\begin{table*}
\centering
\small
\begin{tabular}{
p{\dimexpr 0.2\linewidth-1.2\tabcolsep}
p{\dimexpr 0.09\linewidth-1.2\tabcolsep}
p{\dimexpr 0.1\linewidth-1.2\tabcolsep}
p{\dimexpr 0.09\linewidth-1.2\tabcolsep}
p{\dimexpr 0.09\linewidth-1.2\tabcolsep}
p{\dimexpr 0.09\linewidth-1.2\tabcolsep}
p{\dimexpr 0.1\linewidth-1.2\tabcolsep}
p{\dimexpr 0.09\linewidth-1.2\tabcolsep}} 
\toprule
  & \textbf{LSTM} & \textbf{BiLSTM} & \textbf{GRU} & \textbf{BiGRU} & \textbf{TF-first} & \textbf{HAN-TF} & \textbf{HAN}  \\ 
\hline
F1 on MDL D1 val. $\uparrow$ & 0.966 & 0.965 & 0.961  & 0.959 & 0.898 & \underline{0.976} & \textbf{0.985}  \\
\# of param. per layer $\downarrow$ & 4.72M & 9.45M & \underline{3.54M} & 7.09M  & 3.55M  & 4.14M & \textbf{1.18M}  \\
\bottomrule
\end{tabular}
\caption{\label{tab:encoder_comparison}Comparison results of different encoder layers.  $\uparrow$ denotes that the higher the value is, the better the model is. $\downarrow$ denotes that the lower the value is, the better the model is.}
\end{table*}

\subsection{Encoder benchmarking}
To prove the effectiveness of our HAN encoder, we compare it with six widely used encoders: LSTM, BiLSTM, GRU, BiGRU, TF-first and HAN-TF. HAN encoders are replaced with each of them in our framework. For LSTM and GRU, the final representation of input features is the hidden state of the last token. For BiLSTM and BiGRU, the concatenation of the forward and backward hidden states of the last and first tokens is used. TF-first and HAN-TF are encoders based on Transformer. In TF-first, Transformer is used as an encoder. The hidden state of the first input instance\footnote{We experimented with different fusion methods (e.g., summation, average and linear transformation of concatenated representations) and found that using the hidden state of the first input instance works best.} (a tweet or an MCM) of each user is used by Eq.~\ref{eq:concatenate}. In HAN-TF, we use Transformer to replace the input matrix projection layer (i.e., FNN in Eq.~\ref{eq:key_projection}) of HAN encoders. 

The size of the input and output hidden states of each encoder is set to 768, which is in line with that of BERT-base-uncased. The size of FNNs and the number of heads in Transformer are 768 and 8, respectively. The other hyperparameters remain the same. CNNs are not used as a baseline because they need to be used with other encoders to learn context dependencies~\cite{wang2016dimensional,rhanoui2019cnn}. 

Table~\ref{tab:encoder_comparison} shows F1 scores achieved with the MDL D1 validation set and the number of parameters per encoder layer. HAN encoder achieves better results than all the baseline encoders on our task in terms of F1 score. HAN and HAN-TF outperform LSTM by 1.9\% and 1.0\%, respectively. Although HAN and HAN-TF achieve comparable performance, it is worth noting that using Transformer instead of FNNs in the HAN encoder significantly increases the number of parameters (+2.96M). The number of parameters of the HAN encoder is the smallest among all the encoders. The parameter size of HAN is just a third of that of the second smallest encoder (GRU). 

\begin{figure}[t!]
    \centering
    \includegraphics[width=0.9\linewidth]{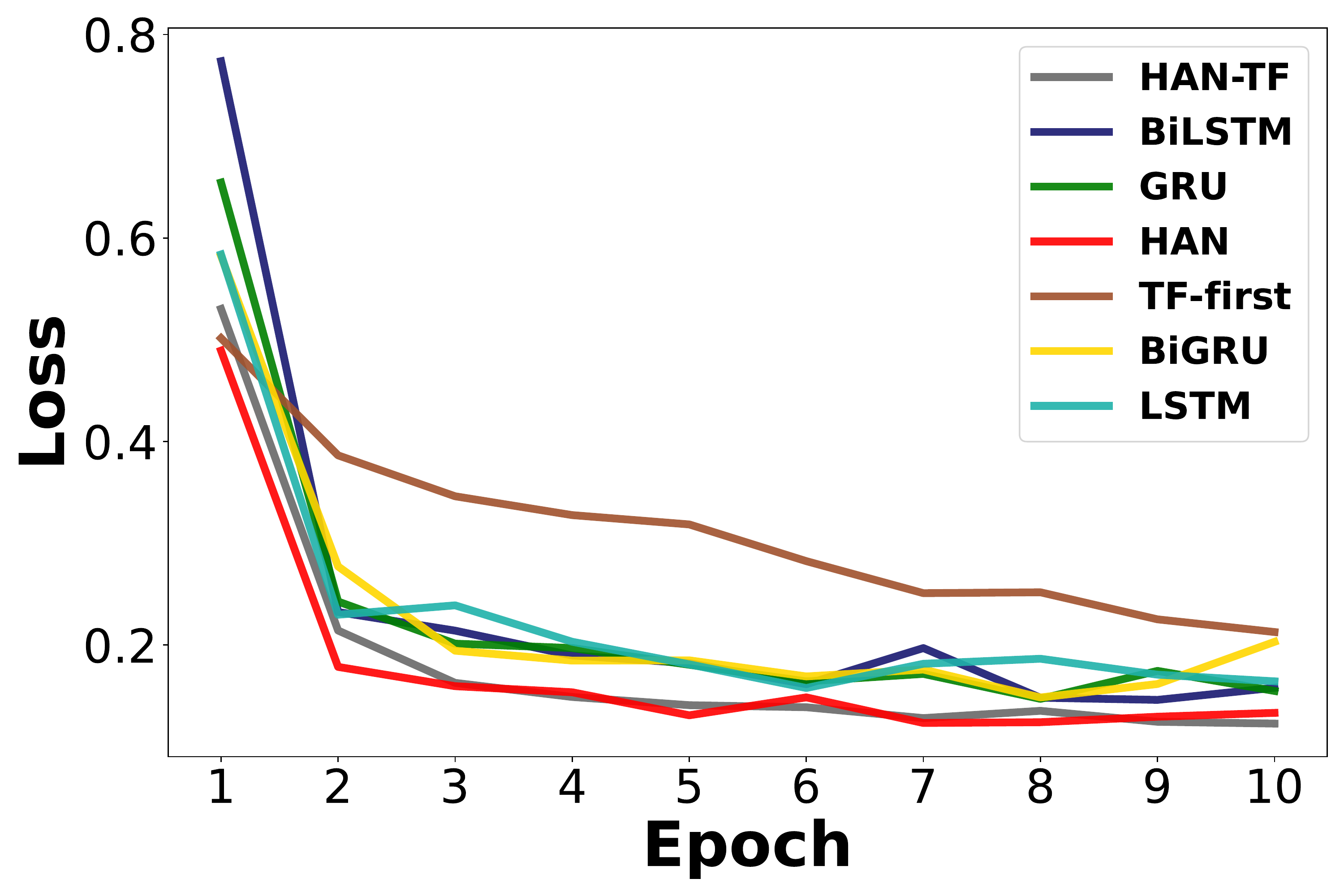}
    \caption{Training loss curves for different encoders obtained using the MDL D1.}
    \label{fg:encoders_lc}
\end{figure} 

Figure~\ref{fg:encoders_lc} shows training loss curves for different encoders plotted using the MDL D1. HAN (the red line) converges faster than the other encoders. Overall, the experiments on different encoders prove that HAN has advantages over the others in terms of effectiveness and efficiency.

\begin{figure}[t!]
\begin{subfigure}{.99\linewidth}
  \centering
  \includegraphics[trim={0 0 7cm 0}, width=\linewidth]{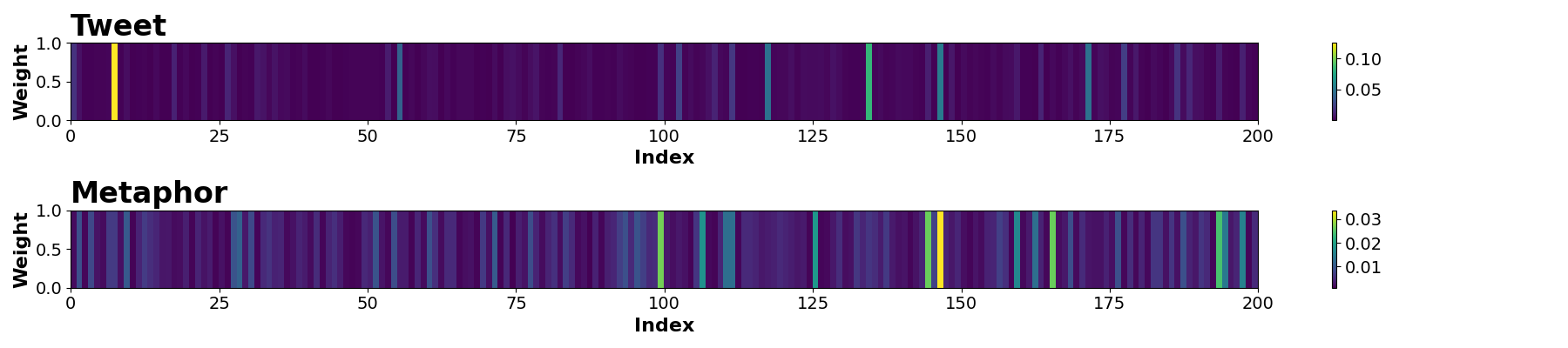}
  \caption{User 1}
\end{subfigure}%

\begin{subfigure}{.99\linewidth}
  \centering
  \includegraphics[trim={0 0 7cm 0}, width=\linewidth]{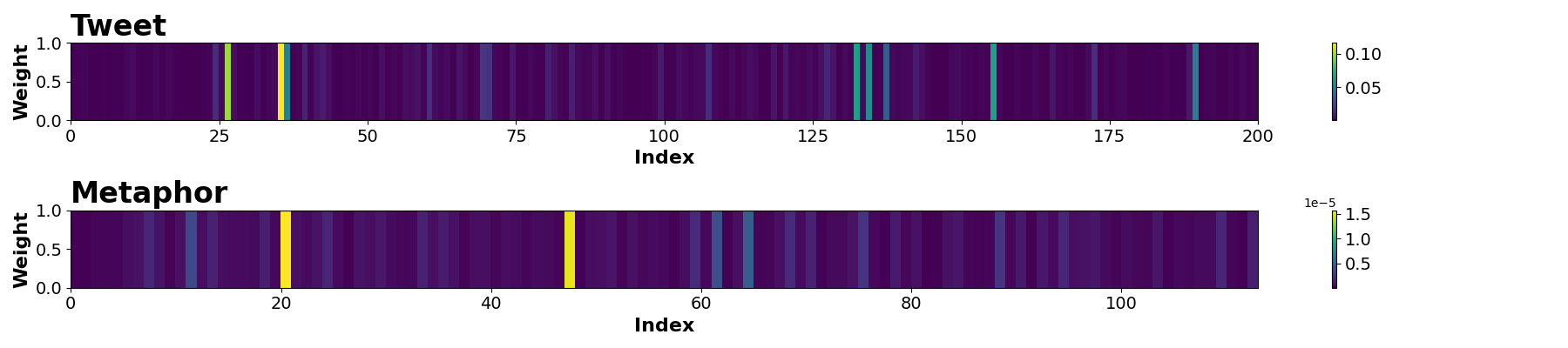}
  \caption{User 2}
\end{subfigure}
\caption{Visualization of attention weights for two depressed users. The lighter the color bar of an instance (tweet or MCM) is, the higher its attention weight is.}
\label{fig:attn_visualisation}
\end{figure}

\begin{table}[htb!]
\centering
\scriptsize
\begin{tabular}{
  p{\dimexpr 0.06\linewidth-1.2\tabcolsep}
  p{\dimexpr 0.45\linewidth-2.0\tabcolsep}
  p{\dimexpr 0.45\linewidth-2.0\tabcolsep}} 
\toprule
\textbf{User} & \textbf{Tweet} & \textbf{Metaphor} \\ \hline
\multirow{11}{*}{1} & 1. I hate how I can't tell if I have allergies or I'm getting sick. & 1. \textsc{level is importance}\\
&2. get better, I love you & 2. \textsc{person is extremity}\\
&3. I'm slightly allergic to cats but I still have them and I don't CARE IF I SNEEZE & 3. \textsc{situation is happening}\\
&4. I'm having a bad night & 4. \textsc{athlete is area}\\
&5. So I'm so nervous for my MAC interview tomorrow but I know I'll do great. Everything will be okay & 5. \textsc{morpheme is extremity}\\ \hline\hline
\multirow{14}{*}{2} & 1. Today is not a good day: Driver, teen shot to death after vehicle hits and kills -year-old & 1. \textsc{concern is state}\\
&2. Autistic th Grader Assaulted by School Cop, Now He is a Convicted Felon & 2. \textsc{position is disappearance}\\
&3. Thank you Father, GM FB! I gotta start taking My butt to bed at night, woke late again & 3. \textsc{level is importance}\\
&4. Cellphone Video Surfaces Showing Moments After Police Shot -Year-Old Boy in the Back  & 4. \textsc{feeling is ill\_health}\\
&5. Freddie Gray dies one week after Baltimore arrest   & 5. \textsc{artifact is support}\\
\bottomrule
  \end{tabular}
  \caption{\label{tab:attention_examples}The top 5 tweets and metaphors, selected based on attention weights, for two example users.}
\end{table}

\subsection{Explainability demonstration}
\label{sec:Explainability demonstration}

Figure~\ref{fig:attn_visualisation} visualizes the attention weights ($w_l$ given by Eq.~\ref{eq:attention weight}) for tweets posted by two example users with depression and MCMs in their tweets. As shown in the figure, HAN can selectively focus on the most important and useful tweets and metaphors by progressively refining feature maps. Attention weights are useful for justifying the decision-making mechanism of HAN because they quantitatively describe how much each tweet and MCM contributes to a predicted label (see Eq.~\ref{eq:attention_sum}). Higher attention weights denote greater utility of tweets and MCMs in detecting depression.

Table~\ref{tab:attention_examples} shows the top 5 tweets and MCMs, ranked according to the attention weights learned during training, for two example users. User~1 tends to use negative expressions to describe personal feelings, state and emotions such as  ``bad'', ``sick'', ``hate'' and ``nervous''. The user also uses positive expressions, such as ``love'', ``I'll do great'' and ``everything will be ok''. However, these positive tweets tend to express self-soothing for negative events. User~2 tends to repost or quote tragic news and add personal comments. The two depressed users show different behaviors on social media, e.g., self-soothing and quoting tragic news.

The listed MCMs in Table~\ref{tab:attention_examples} show that both example users have the same MCM in their tweets, i.e., \textsc{level is importance}. The conceptual projection from \textsc{level} to \textsc{importance} may exacerbate depression because \textsc{level} simply refers to ``a position on a scale of intensity or amount or quality'', whereas \textsc{importance} normally refers to a subjective feeling about ``the worthy of note'' \citep{fellbaum2005wordnet}. 

The imageability of \textsc{importance} may increase stress and anxiety, and thus arouse more depressive feelings \citep{vedhara2003investigation}. For example, there is a tweet saying that ``If a transgender student is bullied, they are put at a \textit{greater} risk of suicide'' posted by a depressed user in the dataset \citep{shen2017depression}. In this tweet, ``\textit{greater}'' is metaphorical. Its contextual meaning refers to a higher risk. ``High'' is one of the manifestations of the target concept \textsc{level}. However, the literal imageability of ``great'' likely refers to the source concept \textsc{importance}, e.g., ``a great work of art'' \citep{fellbaum2005wordnet}. Thus, the metaphorical expression in the example tweet also implies that the ``risk of suicide'' is high and important, which probably increases the subject's nervousness because of their perception about the importance of the risk. We also find that a metaphoric term ``\textit{great}'' is common in the MCM \textsc{level is importance} and its associated tweets posted by depressed users. 

This case study demonstrates that we can further discover common MCMs and metaphorical language patterns among depressed individuals using our proposed model. In general, we argue that HAN is potentially useful for identifying depressed individuals and analyzing different types of such individuals, their cognition and risk factors. 

\section{Ethical Considerations}
This research work was conducted based on a public dataset published by \citet{shen2017depression}. We solely used textual content for concept mapping acquisition, training, and evaluating the model. We did not leverage any information related to user profiles. We oppose the use of our model in any breach of data security, privacy protection, and ethics.

\section{Conclusion}
While most deep learning architectures for depression detection left the impact of different input features on model performance inexplicable, our work attempted to interpret what was going on in the model and justify model predictions. We proposed an attention-based encoder to better understand decision-making process for depression detection. We introduced novel metaphor concept mapping features into our model to investigate how depressed people describe their emotions and experiences. Our extensive experiments and comparative evaluations showed that our model could achieve SOTA performance. An ablation study proved the advantage of utilizing metaphors in depression detection. We argue that a better understanding of metaphors associated with depression can enhance interpretability and help health professionals provide tailored, timely therapy to patients. In future research, we plan to conduct a large-scale study to categorize different characteristics of depression using users' metaphorical and cognitive expressions. 

\section*{Acknowledgments}

We appreciate Rekha Samrajyutha Sajja and Shreyas Nagesh for evaluating metaphor concept mappings. We also appreciate reviewers and Frank Guerin for the constructive comments.

\bibliography{mybib}
\clearpage
\newpage
\appendix

\end{document}